\documentclass[9pt,conference]{IEEEtran}
\IEEEoverridecommandlockouts

\usepackage{amsmath,amssymb,amsfonts}
\usepackage{algorithmic}
\usepackage{graphicx}
\usepackage{textcomp}
\usepackage{stfloats}
\usepackage{url}
\usepackage{verbatim}
\usepackage{graphicx}
\usepackage{cite}
\usepackage{multirow}
\usepackage{lscape,array}
\usepackage{booktabs}
\usepackage{pifont}
\usepackage{ragged2e}
\usepackage{enumitem}
\usepackage{xcolor}
\usepackage[hidelinks]{hyperref}

\def\BibTeX{{\rm B\kern-.05em{\sc i\kern-.025em b}\kern-.08em
    T\kern-.1667em\lower.7ex\hbox{E}\kern-.125emX}}
\begin{document}

\setlength\floatsep{4pt}
\setlength\textfloatsep{6pt}
\setlength\intextsep{6pt}
\setlength\abovecaptionskip{1pt}
\setlength\belowcaptionskip{1pt}
\setlength\abovetopsep{3pt}
\setlength\dblfloatsep{0pt}
\setlength\dbltextfloatsep{6pt}
\aboverulesep=0.25ex 
\belowrulesep=0.5ex 

\title{TS-SUPERB: A Target Speech Processing Benchmark for Speech Self-Supervised Learning Models}


\author{
    \IEEEauthorblockN{ Junyi Peng$^{*,1}$, Takanori Ashihara$^{*,2}$, Marc Delcroix$^2$,  Tsubasa Ochiai$^2$, Oldrich Plchot$^1$, Shoko Araki$^2$,  Jan Černocký$^1$\thanks{The work was supported by European Defense Fund's project No. 101168083 "ARCHER", and Czech Ministry of Interior project No. VJ01010108 "ROZKAZ", Ministry of Education, Youth and Sports of the Czech Republic (MoE) through the OP JAK project "Linguistics, Artificial Intelligence and Language and Speech Technologies: from Research to Applications" (ID:CZ.02.01.01/00/23\_020/0008518). Computing on IT4I supercomputer was supported by the Czech Ministry of Education, Youth and Sports through the e-INFRA CZ (ID:90254).}}
    \IEEEauthorblockA{
  $^1$Brno University of Technology, Czechia, 
  $^2$NTT Corporation, Japan
    }
}

\maketitle

\begin{abstract}
Self-supervised learning (SSL) models have significantly advanced speech processing tasks, and several benchmarks have been proposed to validate their effectiveness. However, previous benchmarks have primarily focused on single-speaker scenarios, with less exploration of target-speaker tasks in noisy, multi-talker conditions---a more challenging yet practical case. In this paper, we introduce the Target-Speaker Speech Processing Universal Performance Benchmark (TS-SUPERB), which includes four widely recognized target-speaker processing tasks that require identifying the target speaker and extracting information from the speech mixture. In our benchmark, the speaker embedding extracted from enrollment speech is used as a clue to condition downstream models. The benchmark result reveals the importance of evaluating SSL models in target speaker scenarios, demonstrating that performance cannot be easily inferred from related single-speaker tasks. Moreover, by using a unified SSL-based target speech encoder, consisting of a speaker encoder and an extractor module, we also investigate joint optimization across TS tasks to leverage mutual information and demonstrate its effectiveness.\footnote{\noindent The code is available at \url{https://github.com/BUTSpeechFIT/TS_SUPERB}. When preparing the code release, we performed some refactoring and refined the recipes to stabilize training. We report updated results in Appendix~\ref{apx:update_results}. \\
* Equal contributors.
}

\end{abstract}%
\section{Introduction}
Many speech processing tasks, including automatic speech recognition (ASR), speaker verification (SV), and speech enhancement (SE), have been significantly advanced by self-supervised learning (SSL) \cite{baevski2020wav2vec, baevski2022data2vec, hsu2021hubert, chen2022wavlm}. This paradigm enables downstream tasks to achieve remarkable performance by exploiting general-purpose features from large SSL models trained on large-scale unlabeled datasets, even with lightweight task-oriented decoders trained on limited amounts of labeled data \cite{li2023parameter, chen2022large, peng2023attention, song2023exploring, hung2022boosting}.

To quantitatively evaluate the capabilities of SSL models across various speech tasks, benchmarks such as the Speech processing Universal PERformance  Benchmark (SUPERB) and its variants have been proposed \cite{yang2021superb, tsai2022superb, shi2023ml, 10389699, 10445941, wu2024emo}. 
SUPERB offers a comprehensive comparison of SSL models across a wide range of downstream speech tasks, such as ASR and SV.
Its downstream tasks consist mostly of single-speaker speech processing tasks.

However, in daily conversational situations, the speech of interest is often interfered with by other speakers (and background noise). To address the complex yet practical conditions, several target-speaker (TS) speech processing tasks, including target speech extraction (TSE), personalized SE (PSE), target-speaker ASR (TS-ASR), and personal voice activity detection (PVAD) \cite{vzmolikova2023TSEoverview,9746962,9383600,ding20_odyssey}, have been developed. For instance, TSE, which aims to extract the TS's speech signal from a mixture, can be used to develop recording devices that focus on the desired speaker for, e.g., hearing aids/hearables or teleconferencing systems applications \cite{veluri2024look}. TS-ASR, which aims to transcribe only the desired user's speech, can be used to develop personalized ASR systems, potentially for use in smart speakers or smart watches \cite{moriya2022streaming}.

Standard TS systems exploit TS's clues, such as speaker embeddings derived from a pre-recorded enrollment utterance, to extract information about the speaker from a multi-talker mixture. 
Overall, TS tasks are more challenging than conventional speech processing tasks because they address two problems simultaneously: identifying the TS and extracting information about the speaker from the speech mixture. We, therefore, characterize the TS tasks, such as TS-ASR, as having a ``dual objective'' in contrast to conventional tasks, such as ASR, that address only one problem, i.e., have a ``single objective,'' because they deal with single-speaker and thus do not require identifying the TS.

Recently, several studies have explored combining TS systems with SSL models \cite{peng2023icassp,10097139,peng2023improving,zhang2023weaklysupervised}. 
Despite the growing interest in TS tasks with SSL models, existing benchmarks have not fully assessed and compared their capabilities in addressing TS tasks.
To advance research in these areas and promote the development of more versatile SSL models, the present study proposes Target Speaker SUPERB (\texttt{TS-SUPERB}). \texttt{TS-SUPERB} includes four widely accepted downstream tasks---TS-ASR, PSE, TSE, and PVAD---with a focus on extracting content (ASR) and fine or coarse acoustic characteristics (SE and VAD, respectively). 

\begin{figure}[t]
    \centering
    \includegraphics[width=0.99\linewidth]{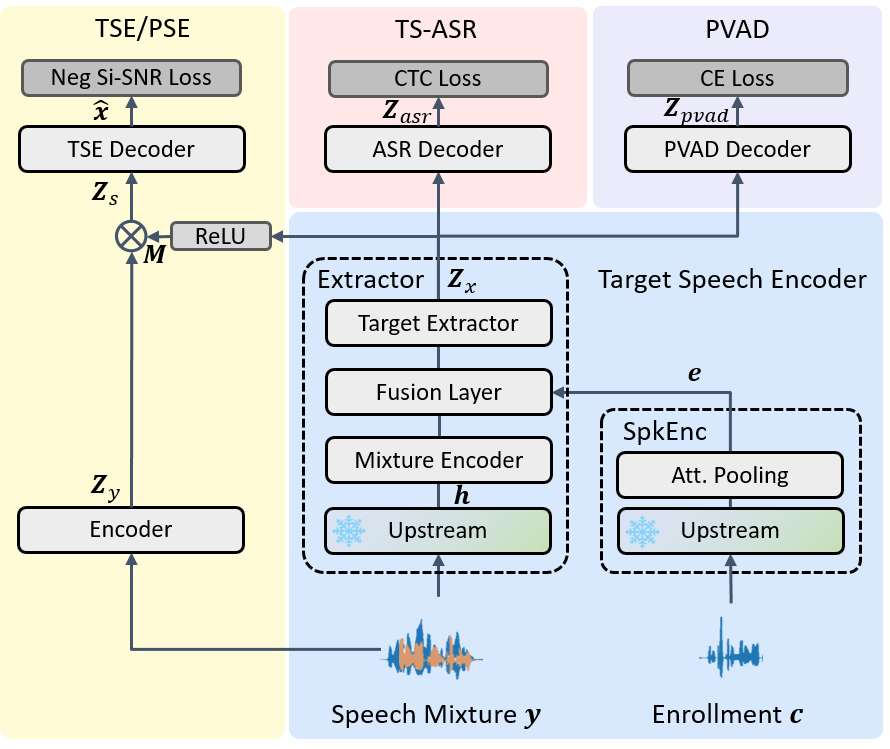}
    \caption{Architecture of proposed TS-SUPERB system. All downstream models use the same architecture for the target speaker encoder, followed by the TSE, PSE, TS-ASR, or PVAD decoder, depending on the task.}
    \label{fig:sys}
\end{figure}

In our benchmark, the downstream models are kept simple, with a structure similar to the related tasks in SUPERB, and consist of two modules: an SSL-based \emph{target speech encoder} and a \emph{task-oriented decoder} as shown in Fig. \ref{fig:sys}. The target speech encoder includes an SSL-based extractor that obtains the TS features, informed by speaker representations derived from enrollment speech via an SSL-based speaker encoder. Since the downstream tasks in \texttt{TS-SUPERB} are related, we can use a similar configuration for the downstream model across all TS tasks, except for the prediction head (or decoder module). The decoder module is task-specific, aligning with the model architecture of related single-speaker tasks in SUPERB.

\texttt{TS-SUPERB} extends the evaluation framework beyond the single-objective SUPERB benchmark by emphasizing various aspects critical for overlapped multi-talker speech processing. 
The key contributions of the proposed \texttt{TS-SUPERB} are:

\textbf{Comprehensive Evaluation:}  The proposed \texttt{TS-SUPERB} provides a standardized framework for evaluating SSL-based TS processing models across four tasks, highlighting SSL's potential in extracting the TS information from mixtures. We will release our benchmark with an evaluation code to encourage further research in SSL-based TS speech processing.

\textbf{Comparative Analysis:} We benchmark seven leading speech SSL models with the proposed \texttt{TS-SUPERB}. Additionally, we analyze the relationship between tasks by computing the layer-wise weight distribution and further
explore correlations between TS tasks and other related tasks (e.g., ASR, SV, and speech separation (Sep)).

\textbf{Shared architecture:} The target speech encoder, which processes both the input mixture and enrollment speech, shares the same architecture across tasks.
By sharing the parameters of the target speech encoder, we can explore jointly training it using multi-task learning across different TS downstream tasks. We demonstrate experimentally that such joint-training can improve performance.


\section{Related work}
Benchmarking SSL models is essential to drive progress in the field. Current speech SSL benchmarks cover a wide variety of downstream tasks \cite{yang2021superb, tsai2022superb, shi2023ml, 10389699, 10445941, wu2024emo, shor20_interspeech, evain21_interspeech, parcollet2024lebenchmark, turian2022hear, conneau22_interspeech, 10597571}, but none include TS speech processing tasks. It is thus difficult to compare the effectiveness of recently proposed SSL schemes for TS tasks. 
For example, there has been an increasing focus on using SSL models for TS speech processing \cite{zhang2023weaklysupervised, huang2023adapting}. In \cite{zhang2023weaklysupervised}, the enrollment speech is incorporated into the pre-training stage as an auxiliary input, achieving superior TS-ASR performance. In \cite{huang2023adapting}, the authors proposed appending speaker embeddings to the Transformer encoder's input within an SSL model and then fine-tuning the entire model for TS-ASR. However, these studies focused on a single TS task. It is thus unclear if the learned representation by these models can also be effective for other TS tasks. 
In this paper, we propose a framework to tackle this issue. Our proposed TS-SUPERB assembled datasets to allow fair comparison between SSL models. We also provide an experimental framework, allowing us to easily benchmark the performance of SSL models for various downstream TS tasks. While the TS tasks are relatively complex compared to single-speaker tasks, our benchmark uses a simple architecture for the downstream model, consisting of a few bidirectional long short-term memory (BLSTM) layers, similar to those used in conventional speech SSL benchmarks, such as SUPERB~\cite{yang2021superb} or SUPERB-SG~\cite{tsai2022superb}.

\section{TS-SUPERB}
In this section, we provide details on the datasets for the multi-talker scenario and the downstream models/tasks within the \texttt{TS-SUPERB} framework, including TSE, PSE, PVAD, and TS-ASR.

\subsection{Datasets}
\label{ssec:datasets}
The training/testing datasets for each task, along with their statistical information, are listed in Table \ref{tab:tasks}.
For the TSE task, we use the Libri2Mix-min clean dataset, which consists of two-speaker mixtures truncated to the duration of the shortest utterance to produce 100\% overlap ratio. 
For PSE and PVAD, we use the Noisy SparseLibri2Mix dataset\cite{cosentino2020librimix}. 
The dataset simulates overlapped speech in conversational audio signals, by mixing speech from LibriSpeech \cite{panayotov2015librispeech} and noise from the WHAM! dataset \cite{wichern2019wham}. Similar to LibriCSS \cite{chen2020continuous}, the overlap ratio in the training set ranges from 0 to 40\%. The training set consists of 24\,000 mixtures, totaling 70 hours. The test dataset contains four overlap conditions (i.e. 0\%, 20\%, 40\%, and 60\%). 
For TS-ASR, we use the Libri2Mix-max clean dataset, which is similar to the Libri2Mix-min except that mixture corresponds to the length of the longest utterance. 

\begin{table}[t]
\caption{Statistics of speech datasets used for the different downstream tasks in \texttt{TS-SUPERB}. }
\centering
\scalebox{0.99}{
\label{tab:tasks}
\begin{tabular}{llc}
\toprule
TASK       & Datasets              & Train/Val/Test (hrs)        \\
\midrule
TSE        & Libri2Mix-min clean\cite{cosentino2020librimix}   &  43.2 / 4.5 / 4.1                                  \\
PSE        & Noisy SparseLibri2Mix\cite{cosentino2020librimix}  & 70.1 / 6.1 / 5.8              \\
PVAD       & Noisy SparseLibri2Mix\cite{cosentino2020librimix} & 70.1 / 6.1 / 5.8                    \\
TS-ASR     & Libri2Mix-max clean\cite{cosentino2020librimix}   &  56.3 / 7.6 / 7.0                                 \\
\midrule
TSE+TS-ASR & Libri2Mix-max clean\cite{cosentino2020librimix}   &  56.3 / 7.6 / 7.0                               \\
PSE+PVAD   & Noisy SparseLibri2Mix\cite{cosentino2020librimix} & 70.1 / 6.1 / 5.8                 \\
\bottomrule
\end{tabular}}
\end{table}

\subsection{Target Speech Encoder}
Figure \ref{fig:sys} presents the unified architecture, emphasizing the integration of pre-trained SSL models. This system aims to isolate the target-speaker information from a mixture $\mathbf{y}$, modeled as $\mathbf{y} = \mathbf{x}+\mathbf{i}$, where $\mathbf{i}$ represents the interference from other speakers and background noise and $\mathbf{x}$ denotes the clean speech. The process utilizes an enrollment utterance $\mathbf{c}$, to identify the TS. The system is structured around two main components: a unified \emph{target speech encoder} that we describe below and a \emph{task-oriented decoder}, which we detail in the next subsections. 

The target speech encoder comprises an extractor and a speaker encoder (SpkEnc), as shown in Fig. \ref{fig:sys}. SpkEnc employs an attentive pooling (i.e., multi-head factorized attention pooling (MHFA) \cite{peng2023attention}) to compute speaker vector $\mathbf{e}$ by processing SSL features based on enrollment speech, $\mathbf{c}$.
Subsequently, the extractor computes the target speech feature, $\mathbf{Z}_x$, from the upstream model's representations, $\mathbf{h}$, and the TS embedding, $\mathbf{e}$, as $\mathbf{Z}_x=\text{Extractor}(\mathbf{h},\mathbf{e})$. Note that, as in SUPERB, $\mathbf{h}$ consists of the weighted sum of the output of the Transformer layers of the SSL model.
Different from other SUPERB tasks, here SSL models are used twice in the downstream model to process the enrollment $\mathbf{c}$ and speech mixture~$\mathbf{y}$. 
Different from using a pre-trained speaker embedding extractor for the SpkEnc, our architectural choice for the SpkEnc allows for future comparisons with recent SSL schemes that handle mixture speech and enrollment speech/embeddings together~\cite{huang2023adapting}.

For the mixture encoder within the target speech encoder, as shown in Fig. \ref{fig:sys}, a single BLSTM layer is used, while the target extractor utilizes two BLSTM layers. The dimension of the BLSTM layers is set to 512, except for the PVAD task, where it is 32. This adjustment is based on preliminary PVAD experiments, which showed no significant performance differences when using models with larger dimensions.
The fusion layer uses broadcast multiplication to combine two features\cite{vzmolikova2023TSEoverview}. The MHFA is configured with four heads and a compression layer with a dimension of 128. 
To align with SUPERB, we keep SSL models frozen by default throughout the training of all tasks (except when specifically mentioned).

\begin{table*}[tb]
\centering
\caption{Comparison of different speech self-supervised models for \texttt{TS-SUPERB} including TSE, PSE, PVAD, and TS-ASR downstream tasks. Additionally, the performance on the corresponding SS, ASR and SV tasks is included for comprehensive evaluation.}
\label{tab:all}
\scalebox{0.9}{
\begin{tabular}{lcccccccccc}
\toprule
\multicolumn{1}{c}{\multirow{2}{*}{Upstream}} & \multicolumn{1}{c}{TSE} & \multicolumn{2}{c}{PSE} & \multicolumn{2}{c}{PVAD} & \multicolumn{2}{c}{TS-ASR (WER)} & \multicolumn{1}{c}{Sep \cite{peng2024probing}} & \multicolumn{1}{c}{ASR \cite{yang2021superb}} & \multicolumn{1}{c}{SV \cite{peng2024probing}} \\
\cmidrule(lr){2-2} \cmidrule(lr){3-4} \cmidrule(lr){5-6} \cmidrule(lr){7-8} \cmidrule(lr){9-9} \cmidrule(lr){10-10} \cmidrule(lr){11-11}
\multicolumn{1}{c}{} & \multicolumn{1}{c}{SI-SDRi$\uparrow$} & \multicolumn{1}{c}{SI-SDRi$\uparrow$} & \multicolumn{1}{c}{FR$\downarrow$} & \multicolumn{1}{c}{mAP$\uparrow$} & \multicolumn{1}{c}{m. tss$\uparrow$} & \multicolumn{1}{c}{w/ LM$\downarrow$} & \multicolumn{1}{c}{w/o LM$\downarrow$} & \multicolumn{1}{c}{SI-SDRi$\uparrow$} & \multicolumn{1}{c}{WER$\downarrow$} & \multicolumn{1}{c}{EER$\downarrow$} \\
\midrule
data2vec Base \cite{baevski2022data2vec} & 9.43 & 10.36 & 4.22 & 0.945 & 0.967 & 33.83 & 40.51 & 9.95 & 4.94 & 3.51 \\
HuBERT Base \cite{hsu2021hubert} & 9.62 & 10.36 & 4.98 & 0.928 & 0.953 & 34.57 & 41.75 & 10.01 & 6.42 & 3.06 \\
WavLM Base \cite{chen2022wavlm} & 10.03 & 11.01 & \textbf{3.58} & 0.951 & 0.971 & 23.72 & 29.13 & 10.80 & 6.21 & 2.71 \\
WavLM Base+ \cite{chen2022wavlm} & \textbf{11.04} & 10.96 & \textbf{3.58} & 0.942 & 0.961 & 23.45 & 29.09 & 11.41 & 5.59 & \textbf{2.03}\\
\midrule
data2vec Large \cite{baevski2022data2vec} & 9.55 & 10.37 & 4.80 & 0.954 & 0.970 & 28.93 & 35.84 & 10.81 & \textbf{3.36} & 2.59\\
HuBERT Large \cite{hsu2021hubert} & 9.03 & 9.76 & 4.25 & 0.954 & 0.973 & 25.10 & 31.88 & 10.95 & 3.62 & 2.94\\
WavLM Large \cite{chen2022wavlm} & 10.47 & \textbf{11.34} & 3.67 & \textbf{0.966} & \textbf{0.980} & \textbf{17.97} & \textbf{22.62} & \textbf{11.87} & 3.44 & 2.30 \\
\bottomrule
\end{tabular}
}
\end{table*}

\subsection{Target Speech Extraction (TSE)}

{\noindent\bf{Task:}} TSE aims to estimate the speech of a TS from a multi-talker mixture \cite{vzmolikova2023TSEoverview}.  It can be used to evaluate both the generative and speaker discrimination capabilities of SSL models.

{\noindent \bf{Architecture:}} In addition to the aforementioned target speech encoder, TSE has a separate encoder that transfers the mixture $\mathbf{y}$ into a sequence of features $\mathbf{Z}_y$. Subsequently, $\mathbf{Z}_x$ are processed through a non-linear activation function, such as ReLU, to compute the target speech mask $\mathbf{M}$, in the feature domain of  $\mathbf{Z}_y$. Finally, the decoder reconstructs the masked features $\mathbf{Z}_s=\mathbf{M} \odot \mathbf{Z}_y$ back into the time domain, resulting in the target speech signal, $\hat{\mathbf{x}}=\text{TSEdecoder}(\mathbf{Z}_s)$. The negative scale-invariant signal-to-noise ratio (SI-SNR) is used as the training objective. Regarding the implementation, we employ Conv1D and DeConv1D for the encoder and decoder, respectively, setting the kernel size to 1024, stride to 320, and the number of filters to 512. 

{\noindent \bf{Evaluation:}} We evaluate TSE performance in terms of scale-invariant signal-to-distortion ratio improvement (SI-SDRi) measured on the Libri2Mix-min, as detailed in Section \ref{ssec:datasets}.


\subsection{Personalized Speech Enhancement (PSE)}
{\noindent\bf{Task:}} Different from TSE, which focuses on processing highly overlapped speech, PSE is more concerned with extracting the TS's voice in conversational scenarios with sparsely overlapped ratios and background noise. The PSE task probes the generative and speaker discrimination capabilities of SSL models with noisy multi-talker inputs.

{\noindent \bf{Architecture:}} The downstream model architecture is the same as for~TSE.

{\noindent \bf{Evaluation:}} 
We use the SparseLibri2Mix dataset, described in Section \ref{ssec:datasets}, to evaluate the PSE task.
The final performance is measured in terms of SI-SDRi, perceptual evaluation of speech quality (PESQ), short-time objective intelligibility (STOI), and Failure rate (FR) \cite{delcroix2022listen}, where FR is defined as the percentage of test samples with SDRi below 1 dB. 

\subsection{Personalized VAD (PVAD)}

{\bf{Task:}} PVAD is designed to detect the voice activity of a TS in a mixture at frame level\cite{ding20_odyssey}. Unlike VAD, which simply detects the presence of any speech or not, PVAD aims to detect the TS's voice in a multi-talker recording. 

{\noindent \bf{Architecture:}} The PVAD decoder employs a classification layer (i.e., a linear layer with softmax activation function) to predict frame-level labels with cross-entropy loss between actual labels. Each utterance is annotated with frame-level labels covering three categories, i.e., target speaker speech (tss), non-target speaker speech (ntss), and none-speech (ns), such as noise and silence.

{\noindent \bf{Evaluation:}} The dataset used for PVAD aligns with that for PSE, i.e., noisy SparseLibri2Mix. We measure performance using mean average precision (mAP) and mAP of target speaker speech (m. tss)~\cite{ding20_odyssey}. 

\subsection{Target-speaker ASR (TS-ASR)}

{\bf{Task:}} TS-ASR transcribes target-speaker speech into words given the multi-talker mixture. This task serves to evaluate both the linguistic and speaker discrimination capabilities of SSL models.

{\noindent \bf{Architecture:}} The decoder within TS-ASR employs a single BLSTM layer with 512 dimensions. It estimates the probability of characters using the connectionist temporal classification (CTC) loss. The trained model is decoded with the official LibriSpeech 4-gram LM, making use of KenLM \cite{heafield2011kenlm} and the Flashlight toolkit \cite{pratap2019wav2letter}.

{\noindent \bf{Evaluation:}} The performance of TS-ASR is evaluated on the Libri2Mix-max dataset. The evaluation metric is the word error rate~(WER).

\subsection{Multi-task Learning}
All \texttt{TS-SUPERB} tasks share the same architecture for the target speech encoder. To leverage the potential mutual information between different tasks and further enhance the learned features, we explore a multi-task learning framework for joint optimization across tasks. Here we investigate multi-task learning between two tasks (TSE+TS-ASR and PSE+PVAD). The optimization objective can be formulated as $\mathcal{L}=\alpha \mathcal{L}_{i}+(1-\alpha) \mathcal{L}_{j}$, where $0 \leq \alpha \leq 1$ denotes the task weight. $\mathcal{L}_{i}$ and  $\mathcal{L}_{j}$ are the loss functions of the two tasks, i.e., $(i,j)\in\{(\text{TSE},\text{TS-ASR}),(\text{PSE},\text{PVAD})\}$.



\section{Results and Analysis}

\subsection{Performance comparison of various SSL models}

The results of evaluating upstream models on \texttt{TS-SUPERB} are presented in Table \ref{tab:all}. Additionally, to explore the correlations between tasks, we also include results from related ASR, SV, and Sep tasks following implementation in S3PRL, within the same table. Specifically, for PSE, we report the averaged SI-SDRi and FR metrics across four different overlap ratio conditions. 
Overall, no single SSL model consistently leads across all tasks.
Meanwhile, the WavLM Base+ and Large models exhibit competitive performance in TS tasks, likely due to their use of a training data augmentation strategy, which leverages mixed utterances from
multiple speakers, making
them well-suited for the multi-speaker and noisy conditions of \texttt{TS-SUPERB}.

In detail, WavLM Base shows the best performance in the PVAD task, followed by data2vec Base and WavLM Base+ among all BASE models, though the lead is marginal. For TS-ASR, the Large model significantly outperforms the Base model, attributed to its stronger modeling of phoneme-related hidden units during pre-training. 
In denoising tasks (PSE and TSE), WavLM models obviously outperform others, possibly because of denoising strategies incorporated during their pre-training stage as data augmentation. 

Note that this paper aims to compare SSL models over various TS tasks and not to achieve top performance on these tasks. However, for reference, we compared the results of Table \ref{tab:all} with prior TSE and TS-ASR works.
SSL-based models demonstrate comparable performance with state-of-the-art systems trained from scratch like TD-SpeakerBeam \cite{delcroix2020improving}, which achieves SI-SDRi of  13.03 dB and 10.71 dB on the TSE and PSE tasks, respectively, suggesting the promising applications of SSL models in denoising. However, the performance of the TS-ASR systems is significantly behind that reported in \cite{10097139}, which employed a pre-trained speaker encoder, conditioned the pre-trained SSL model on the TS embeddings, and performed full fine-tuning of the SSL model, resulting in a WER of 12.32 \% with a different configuration for enrollment speech. 
This indicates the potential for future improvement and investigation.

\begin{figure}[tb]
    \centering
    \includegraphics[width=0.75\linewidth]{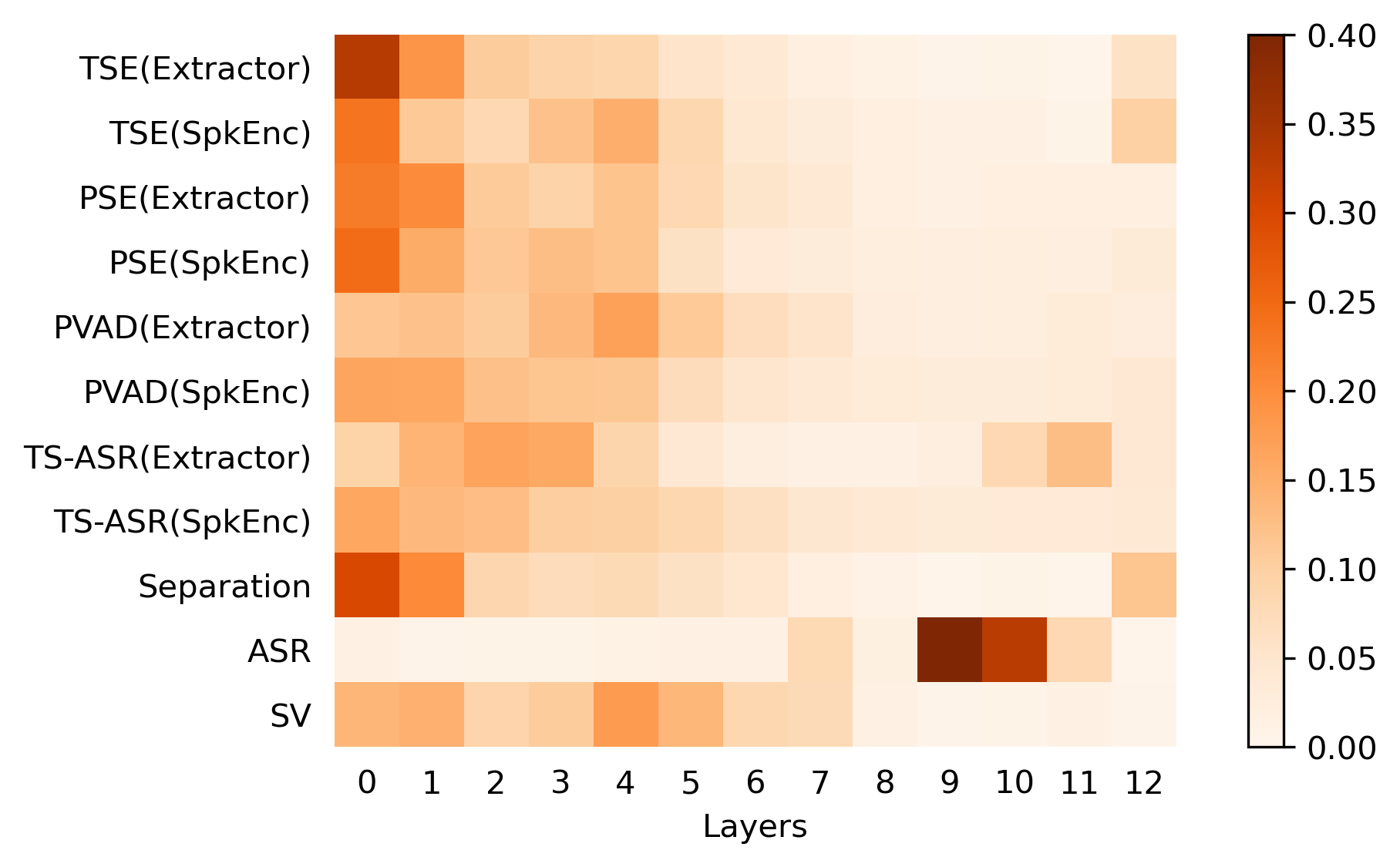}
    \caption{The weight distribution of SSL layers. Note that the 0th layer denotes the input of the 1st Transformer encoder layer.}
    \label{fig:weights}
\end{figure}

\begin{figure}[tb]
    \centering
    \includegraphics[width=0.55\linewidth]{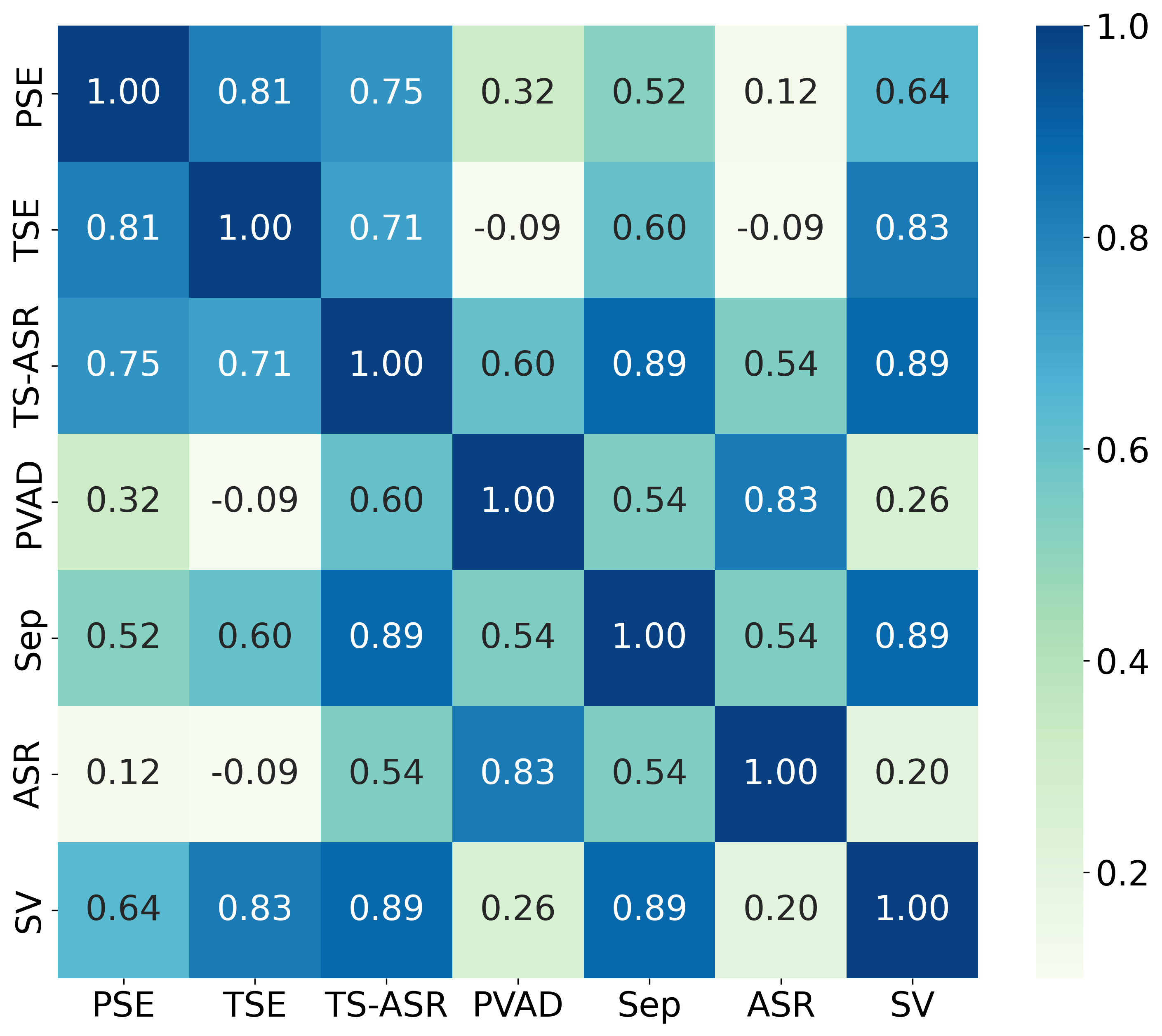}
    \caption{The Spearman's rank correlation between tasks. }
    \label{fig:corr}
\end{figure}

\subsection{Layer-wise and task correlation analysis}

The layer weights distributions of the WavLM Base+ model across various tasks are shown in Fig.~\ref{fig:weights}. In TS tasks, the contributions of the bottom layers in both the extractor and the SpkEnc are more significant than for tasks such as SV, Sep, and ASR. SpkEnc's weight distribution resembles that in SV, indicating the bottom layer of the Base model contains rich speaker-related information.  In addition, both the TS-ASR extractor and ASR show a peak at the 10th layer, suggesting that semantic information is encoded in the top layers. However, unlike ASR, TS-ASR also requires information from the lower layers to identify the TS in the mixture.

Furthermore, we investigate the correlations between tasks within \texttt{TS-SUPERB} and related tasks to verify if task relationships confirm existing hypotheses in speech processing. 
After converting all metrics to a higher-is-better format, we then calculate Spearman's correlation coefficients among all tasks \cite{spearman1904proof}. As illustrated in Fig.~\ref{fig:corr}, all \texttt{TS-SUPERB} tasks exhibit a strong correlation with Sep. One possible reason is that these tasks share a similar goal: extracting specific speaker information from mixed speech signals. SV shows a high correlation with the TS tasks, except for PVAD, suggesting that PVAD does not require strict speaker representations. TSE, PSE, and TS-ASR show a high correlation, whereas PVAD has a weaker correlation with TSE and PSE but a stronger connection with TS-ASR and ASR. One possible explanation is that PVAD, TS-ASR, and ASR focus on frame-level feature processing, which does not require the fine-grained temporal resolution needed by TSE and PSE.
It is observed that TS-ASR has a surprisingly weak correlation with ASR. This could be attributed to the additional requirement of TS-ASR to separate the mixture speech, which complicates the task using the simple ASR decoder module, making it challenging to simultaneously deal with both tasks effectively.

\begin{table}[tb]
\centering
\caption{Investigation of joint-training of TSE and TS-ASR on the Libri2Mix-max clean data.}
\label{tab:e2e_asr}
\scalebox{0.85}{
\begin{tabular}{lcccc}
\toprule
Model  & SI-SDRi$\uparrow$  & STOI(\%)$\uparrow$ & PESQ$\uparrow$ & WER$\downarrow$ \\
\midrule
TSE+TS-ASR  & \textbf{11.74} & \textbf{90.60} & \textbf{2.14} & 33.26 \\
- Only TSE  & 11.16 & 89.40 & 1.97 &  -  \\
- Only TS-ASR  & - & - & - &  \textbf{29.09} \\
\midrule
TSE+TS-ASR [FT]  & 12.70 & 91.66 & 2.24 & 26.29 \\
\bottomrule
\end{tabular}
}
\end{table}

\begin{table}[t]
\centering
\caption{Performance on noisy SparseLibri2Mix. All results are averages of four different overlap ratios (i.e. 0\%, 20\%, 40\%, 60\%) testing data. \#Params denotes the trainable parameters.
}
\label{tab:pvad_pse}
\scalebox{0.9}{
\begin{tabular}{lccc}
\toprule
Model & \#Params  & SI-SDRi$\uparrow$ & mAP$\uparrow$ \\
\midrule
PSE+PVAD  & 7.3 M & \textbf{11.09} & \textbf{0.962}  \\
- Only PSE  & 7.3 M & 10.96 & -    \\
- Only PVAD  & 0.7 M & - & 0.942    \\
\midrule
PSE+PVAD [FT] & 101M & 11.49 & 0.961\\
\bottomrule
\end{tabular}
}
\end{table}

\subsection{Analysis of multi-task learning and fine-tuning}
To investigate how the target speech processing tasks complement each other, we evaluate the performance of joint-training TSE and TS-ASR in Table \ref{tab:e2e_asr} and PSE with PVAD in Table \ref{tab:pvad_pse}, respectively. The first experiments are conducted using Libri2Mix-max, while the second utilizes noisy SparseLibri2Mix. The $\alpha$ is set to $0.5$ for all experiments. WavLM Base+ was used for the upstream model.

When jointly training TSE and TS-ASR tasks (Table \ref{tab:e2e_asr}), the system achieves improved TSE performance in terms of SI-SDRi, STOI, and PESQ, compared to TSE alone but 
worse WER than standalone TS-ASR. 
The fine-tuning of the whole system results in further improvement in all metrics.

In Table \ref{tab:pvad_pse}, the jointly optimized model improves in both PSE and PVAD performance. This indicates that speaker detection and denoising tasks can mutually benefit each other. Notably, training with an unfrozen SSL model leads to further improvements, especially for the TSE+TS-ASR task.

\section{Conclusion}
In this paper, we propose a new benchmark, named \texttt{TS-SUPERB}, for target-speaker speech processing, comprising four new tasks: TSE, PSE, TS-ASR, and PVAD. We further explore the multi-task learning within those tasks to utilize their mutual information. Through the comprehensive experiments with seven SSL models evaluated on Libr2Mix and noisy SparseLibri2Mix datasets, our results demonstrate the uniqueness of target speech processing tasks as their performances cannot be simply deduced from SV, ASR, and Sep tasks alone. 

\appendices
\section{Experimental Results from the Released Codebase}
\label{apx:update_results}
In this appendix, we report the results obtained using the refactored code after the release of the paper.
The results for TSE, PSE, TS-ASR, and PVAD, shown in Tables~\ref{atab:tse},~\ref{atab:pse},~\ref{atab:tsasr}, and~\ref{atab:pvad}, respectively, slightly differ from those in Table~\ref{tab:all}.
Moreover, we provide results for the PSE and PVAD experiments with different overlap ratios.

{
\setlength\aboverulesep{0.4ex}
\setlength\belowrulesep{0.65ex}
\setlength\abovecaptionskip{1pt}
\setlength\belowcaptionskip{0pt}

\begin{table}[h]
\centering
\caption{Target Speech Extraction (TSE) results}
\label{atab:tse}
\begin{tabular}{lccc}
\toprule
\multicolumn{1}{c}{\multirow{2}{*}{Upstream}} & \multicolumn{3}{c}{TSE} \\ \cmidrule(lr){2-4}
\multicolumn{1}{c}{} & \multicolumn{1}{c}{SI-SDRi$\uparrow$} & \multicolumn{1}{c}{STOI$\uparrow$} & \multicolumn{1}{c}{PESQ$\uparrow$} \\
\midrule
HuBERT Base \cite{hsu2021hubert} & 9.64 & 87.30 & 1.744 \\
WavLM Base \cite{chen2022wavlm} & 10.26 & 88.40 & 1.858 \\
WavLM Base+ \cite{chen2022wavlm} & \textbf{10.69} & \textbf{89.00} & \textbf{1.915} \\
\bottomrule
\end{tabular}
\end{table}

\begin{table}[h]
\centering
\caption{Personalized Speech Enhancement (PSE) results for each overlap ratio (0, 20, 40, and 60\%) and metric (SI-SDRi$\uparrow$, STOI$\uparrow$, and PESQ$\uparrow$)}
\label{atab:pse}
\begin{tabular}{lccccc}
\toprule
\multicolumn{1}{c}{\multirow{2}{*}{Upstream}} & \multicolumn{5}{c}{PSE (SI-SDRi$\uparrow$)} \\ \cmidrule(lr){2-6}
\multicolumn{1}{c}{} & \multicolumn{1}{c}{0\%} & \multicolumn{1}{c}{20\%} & \multicolumn{1}{c}{40\%} & \multicolumn{1}{c}{60\%} & \multicolumn{1}{c}{Avg.} \\
\midrule
HuBERT Base \cite{hsu2021hubert} & 10.65 & 8.89 & 7.85 & 7.08 & 8.61 \\
WavLM Base \cite{chen2022wavlm} & 11.03 & 10.08 & 8.85 & 7.84 & 9.65 \\
WavLM Base+ \cite{chen2022wavlm} & \textbf{11.94} & \textbf{10.55} & \textbf{9.22} & \textbf{8.33} & \textbf{10.01} \\
\bottomrule
\vspace{3pt}
\end{tabular}
\begin{tabular}{lccccc}
\toprule
\multicolumn{1}{c}{\multirow{2}{*}{Upstream}} & \multicolumn{5}{c}{PSE (STOI$\uparrow$)} \\ \cmidrule(lr){2-6}
\multicolumn{1}{c}{} & \multicolumn{1}{c}{0\%} & \multicolumn{1}{c}{20\%} & \multicolumn{1}{c}{40\%} & \multicolumn{1}{c}{60\%} & \multicolumn{1}{c}{Avg.} \\
\midrule
HuBERT Base \cite{hsu2021hubert} & 86.10 & 81.20 & 77.50 & 74.90 & 79.92 \\
WavLM Base \cite{chen2022wavlm} & 87.30 & 82.90 & 79.40 & 76.70 & 81.57 \\
WavLM Base+ \cite{chen2022wavlm} & \textbf{87.90} & \textbf{83.90} & \textbf{80.80} & \textbf{78.10} & \textbf{82.67} \\
\bottomrule
\vspace{3pt}
\end{tabular}
\begin{tabular}{lccccc}
\toprule
\multicolumn{1}{c}{\multirow{2}{*}{Upstream}} & \multicolumn{5}{c}{PSE (PESQ$\uparrow$)} \\ \cmidrule(lr){2-6}
\multicolumn{1}{c}{} & \multicolumn{1}{c}{0\%} & \multicolumn{1}{c}{20\%} & \multicolumn{1}{c}{40\%} & \multicolumn{1}{c}{60\%} & \multicolumn{1}{c}{Avg.} \\
\midrule
HuBERT Base \cite{hsu2021hubert} & 1.591 & 1.379 & 1.284 & 1.242 & 1.374 \\
WavLM Base \cite{chen2022wavlm} & 1.688 & 1.443 & 1.337 & 1.202 & 1.437 \\
WavLM Base+ \cite{chen2022wavlm} & \textbf{1.737} & \textbf{1.480} & \textbf{1.372} & \textbf{1.313} & \textbf{1.475} \\
\bottomrule
\end{tabular}
\end{table}

\begin{table}[h]
\centering
\caption{Target-speaker automatic speech recognition (TS-ASR) results}
\label{atab:tsasr}
\begin{tabular}{lcc}
\toprule
\multicolumn{1}{c}{\multirow{2}{*}{Upstream}} & \multicolumn{2}{c}{TS-ASR} \\ \cmidrule(lr){2-3}
\multicolumn{1}{c}{} & \multicolumn{1}{c}{w/ LM$\downarrow$} & \multicolumn{1}{c}{w/o LM$\downarrow$} \\
\midrule
HuBERT Base \cite{hsu2021hubert} & 30.52 & 36.86 \\
WavLM Base \cite{chen2022wavlm} & 22.68 & 27.82 \\
WavLM Base+ \cite{chen2022wavlm} & \textbf{20.06} & \textbf{24.75}\\
\bottomrule
\end{tabular}
\end{table}

\begin{table}[h]
\centering
\caption{Personalized voice activity detection (PVAD) results for each overlap ratio (0, 20, 40, and 60\%)}
\label{atab:pvad}
\begin{tabular}{lccccc}
\toprule
\multicolumn{1}{c}{\multirow{2}{*}{Upstream}} & \multicolumn{5}{c}{PVAD (mAP$\uparrow$)} \\ \cmidrule(lr){2-6}
\multicolumn{1}{c}{} & \multicolumn{1}{c}{0\%} & \multicolumn{1}{c}{20\%} & \multicolumn{1}{c}{40\%} & \multicolumn{1}{c}{60\%} & \multicolumn{1}{c}{Avg.} \\
\midrule
HuBERT Base \cite{hsu2021hubert} & \textbf{94.25} & 94.75 & 94.74 & 94.66 & 94.60 \\
WavLM Base \cite{chen2022wavlm} & 94.09 & 94.57 & 94.78 & 94.35 & 94.40 \\
WavLM Base+ \cite{chen2022wavlm} & 93.96 & \textbf{95.00} & \textbf{95.52} & \textbf{95.88} & \textbf{95.00} \\
\bottomrule
\end{tabular}
\end{table}

}

\newpage
\bibliographystyle{IEEEtran}
\bibliography{mybib}
\end{document}